\documentclass{article}
\usepackage{spconf,amsmath,graphicx,amssymb,multirow,setspace}

\title{Gaussian Kernel-based Cross Modal Network for \\ Spatio-Temporal Video Grounding}
\name{Zeyu Xiong$^{1}$ \qquad Daizong Liu$^{2}$ \qquad Pan Zhou$^{1}\sthanks{Corresponding author.}$}
\address{$^{1}$
The Hubei Engineering Research Center on Big Data Security, \\School of Cyber Science and Engineering, Huazhong University of Science and Technology  \\
$^{2}$ Wangxuan Institute of Computer Technology, Peking University\\
zeyuxiong@hust.edu.cn, dzliu@stu.pku.edu.cn, panzhou@hust.edu.cn
}

\begin{document}
%
\maketitle
\normalsize
\begin{abstract}
Spatial-Temporal Video Grounding (STVG) is a challenging task which aims to localize the spatio-temporal tube of the interested object semantically according to a natural language query. Most previous works not only severely rely on the anchor boxes extracted by Faster R-CNN, but also simply regard the video as a series of individual frames, thus lacking their temporal modeling. Instead, in this paper, we are the first to propose an anchor-free framework for STVG, called Gaussian Kernel-based Cross Modal Network (GKCMN). Specifically, we utilize the learned Gaussian Kernel-based heatmaps of each video frame to locate the query-related object. A mixed serial and parallel connection network is further developed to leverage both spatial and temporal relations among frames for better grounding. Experimental results on VidSTG dataset demonstrate the effectiveness of our proposed GKCMN.
\end{abstract}

\vspace{-4pt}
\section{Introduction}
\begin{figure}[t]
\centering

\includegraphics[width=1.0\columnwidth]{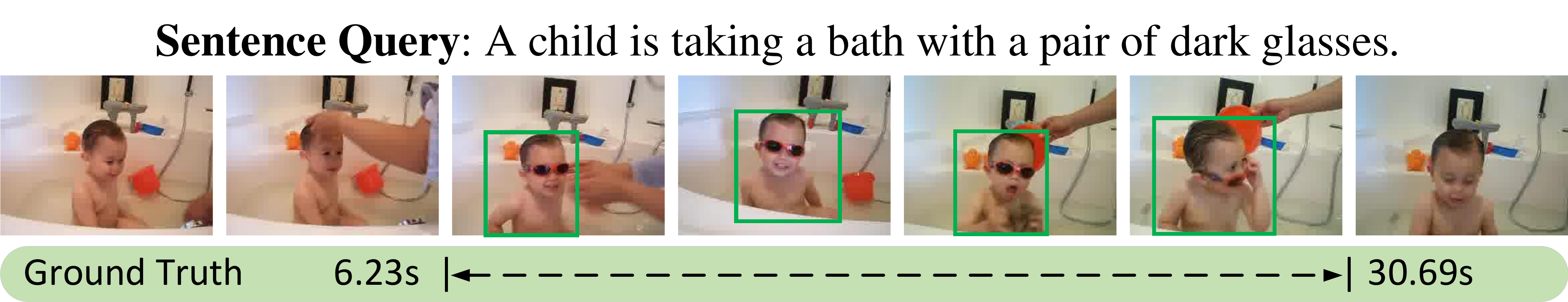}

\caption{An example of spatio-temporal video grounding task.}
\label{fig1}
\vspace{-10pt}
\end{figure}

\vspace{-4pt}
\label{sec:intro}
Video grounding with natural language is a fundamental but challenging problem due to its vast potential applications in visual-language understanding. Generally, it could be categorized into three different classes: spatial grounding \cite{hu2016natural,yu2018mattnet,yang2019cross,yang2019dynamic}, temporal grounding \cite{gao2017tall,anne2017localizing,liu2020jointly,liu2021context} and spatio-temporal grounding \cite{zhang2020does}.
Among them, Spatial-Temporal Video Grounding (STVG) is substantially more challenging as it needs to not only model the complicated multi-modal interactions for semantics alignment, but also retrieve both spatial location and temporal duration of the target activity. As shown in Fig. \ref{fig1}, STVG aims to localize the spatio-temporal tube of the queried object according to the given textual description.

Most previous spatial or temporal video grounding technologies \cite{zhou2018weakly,vasudevan2018object,yamaguchi2017spatio} are designed to tackle the grounding problems by directly detecting the foreground objects of each video frame for objects correlation learning \cite{vasudevan2018object,yamaguchi2017spatio}, or by regressing the temporal segment boundary in the video \cite{zhou2018weakly}. One most recent work for STVG task \cite{zhang2020does} tackles the grounding problem in a more general way, which is able to ground the spatio-temporal tubes of the queried object in untrimmed videos. 
However, all these grounding methods suffer the following drawbacks: (1) They severely rely on the detection quality of the detection model. Besides, they generally pre-generate the proposal regions with the detected anchor boxes, leading to the time-consuming computation.
(2) They typically treat the video frames individually without considering the temporal correlation between the consecutive frames.

To alleviate the above issues, in this paper, we introduce the first anchor-free framework for STVG task, called Gaussian Kernel-based Cross Modal Network (GKCMN). 
Specifically, for video clip encoding, we extract their spatio-temporal feature maps from the last convolutional layer of I3D \cite{carreira2017quo}. For sentence query encoding, we first extract the word-level features individually by the Glove \cite{pennington2014glove} and then embed their sequential information via a BiGRU. Both visual and sentence semantics are subsequently aligned by using a Cross-model Interaction module. After that, a Mixed Serial and Parallel Connection Network is further deployed to learn both spatial non-local information and temporal modeling of the multi-modal representation.
As for spatial locating, we learn Gaussian heatmaps for indicating the object position in each frame by utilizing a Gaussian kernel to mark the position of multi-modal semantics and deeming every pixel in annotation boxes as a boundary regression sample during the training process. At last, to accurately localize the start and end timestamps of the queried objects, we calculate the confidence score of the potential temporal candidates and develop a Boundary Regression Head to rectify the offsets of the highest score temporal tube. 

\begin{figure*}[t]
\centering
\includegraphics[width=1.0\textwidth]{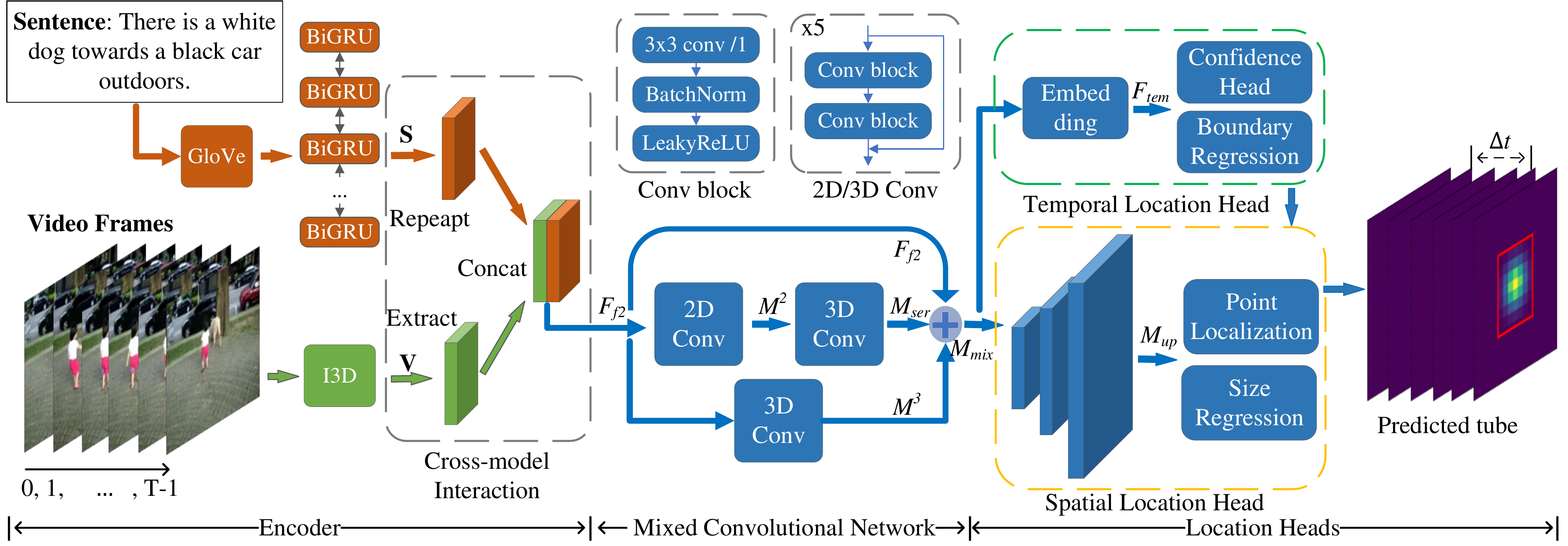}
\vspace{-15pt}
\caption{The overall architecture of our proposed GKCMN.}
\label{fig2}
\vspace{-8pt}
\end{figure*}
Our main contributions are summarized as:
(1) We propose the first anchor-free model GKCMN for spatio-temporal video grounding, which utilizes Gaussian kernels to highlight the foremost region of target semantics;
(2) We design a mixed convolutional network to capture both temporal and spatial information;
(3) We demonstrate the effectiveness of the GKCMN model by evaluating on the VidSTG dataset.

\vspace{-4pt}
\section{Method}
\vspace{-4pt}
Given an untrimmed video $V$ and a sentence $S$, the STVG task aims to retrieve a spatio-temporal tube $U$ in video $V$, which corresponds to the semantic of sentence $S$.
The framework of our proposed GKCMN is illustrated in Fig. \ref{fig2}.

\vspace{-4pt}
\subsection{Encoder}
\vspace{-3pt}
\noindent\textbf{Video Encoder.} To encode the video frames, we utilize a pre-trained I3D network \cite{carreira2017quo} to extract its last convolutional layer features as $\mathbf{V} =\left\{ \mathbf {v}_t \right\} ^{T-1}_{t=0}$, where $T$ denotes the frame number, $\mathbf {v}_t \in\mathbb{R} ^ {d \times \frac{H}{r_h} \times \frac{W}{r_w}}$ represents the $t$-th frame 2D feature, $d$ is visual feature dimension, $H$ and $W$ are the height and width of input frames, and $r_h, r_w$ are scaling factors.

\noindent\textbf{Sentence Encoder.} For sentence encoding, we first extract word-level features via the Glove \cite{pennington2014glove} embedding, and then employ a self-attention module \cite{vaswani2017attention} to capture the self-dependencies among words. We further utilize a Bi-GRU \cite{chung2014empirical} to learn their sequential features and denote the final sentence feature as $\mathbf{S} = \left\{ \mathbf {s}_n \right\} ^{N-1}_{n=0} $, where $\mathbf {s}_n \in\mathbb{R} ^ {D}$  represents the $n$-th word feature, and $D$ is word feature dimension.

\noindent\textbf{Cross-model Interaction.} We first repeat the textual tensor $\mathbf{S}_r = {\rm repeat}(\mathbf{S})$ to the same shape with visual tensor, and obtain the multi-modal matrix by fusing $\mathbf{V}$ with $\mathbf{S}_r$. Next, the cross-modal interacted feature $F_{f2}$ is obtained by:
\begin{equation}
\begin{aligned}
&F_{f1} = g(\mathbf{V}W_{v1}) \cdot g(\mathbf{S}_{r}W_{s1}),\\
&F_{f2} = {\rm concat}(g(F_{f1}W_{f2}), g(\mathbf{V}W_{v2})),
\end{aligned}
\end{equation}
where $g(\cdot)$ is the non-linear activation function, $W_{v1}$, $W_{s1}$, $W_{f2}$ and $W_{v2}$ are learnable parameters. In following process, we utilize $F_{f2} \in \mathbb{R}^{D^{\prime}\times T \times  \frac{H}{r_h} \times \frac{W}{r_w}}$ as the input fusion feature to extract spatio-temporal relationships. 
\vspace{-3pt}
\subsection{The Mixed Convolutional Network}
\vspace{-3pt}
We construct spatio-temporal relationships in terms of depth and width by the serial connection network and the parallel connection network, respectively.

\noindent\textbf{Serial Connection Network.} For any 3D signal $F_{f2}$, we reshape it into the 2D batches and learn the spatial characteristics of each frame by 2D convolutional blocks. Next, we use 3D convolution to learn the timing sequential relationship between each activity. Therefore, the output of the serial connection network ${\cal M}_{ser}$ can be formulated as:
\begin{equation}
{\cal M}_{ser} =  {\cal K}^3 \otimes {\rm reshape}^{\prime}({\cal K}^2 \otimes {\rm reshape}(F_{f2})),
\end{equation}
where $\otimes$ represents the convolution operation, ${\cal K}^i$ means the $i$-dimension kernel.

\noindent\textbf{Parallel Connection Network.} Similarly, we utilize parallel structures to learn and fuse stationary and temporal information via 2D and 3D CNN blocks.
\begin{equation}
{\cal M}_{par} =  {\cal K}^3 \otimes F_{f2} + {\rm reshape}^{\prime}({\cal K}^2 \otimes {\rm reshape}(F_{f2})).
\end{equation}

\noindent\textbf{Mixed Convolutional Network.} Our Mixed Convolutional Network combines series and parallel connections with a residual structure.
As shown in Fig. \ref{fig2}, we fuse the original feature maps $F_{f2}$, serial features ${\cal M}_{ser}$ and temporal residual features ${\cal M}^3$ via a triple-parallel structure. Then we have:
\begin{equation}
{\cal M}_{mix} =  {\cal K}^3 \otimes F_{f2} + {\cal M}_{ser} + F_{f2}.
\end{equation}

\subsection{Spatial Location Head}
\vspace{-3pt}
We construct a Gaussian kernel-based spatial location head to predict bounding boxes of queried object. Firstly, we up-sample ${\cal M}_{mix}$ to obtain ${\cal M}_{up} \in \mathbb{R} ^ { {D}^{\prime} \times T \times {L} \times L} $ for spatial localization scaling, where $L$ is the feature map size.

\begin{figure}[t]
\centering
\includegraphics[width=0.85\columnwidth]{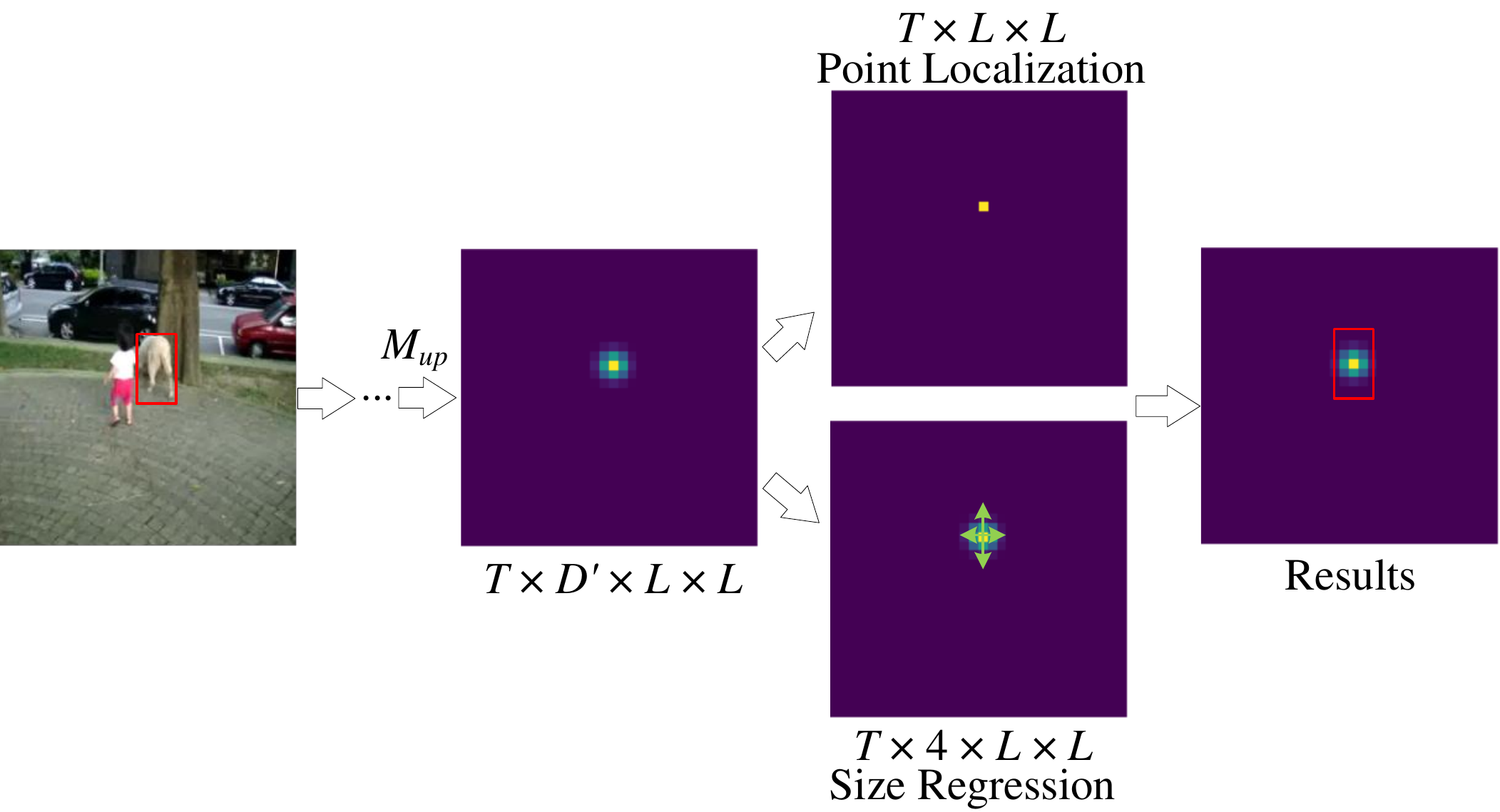}
\vspace{-3pt}
\caption{Illustration of Gaussian kernel-based grounding.}
\label{figspa}
\vspace{-8pt}
\end{figure}

\noindent\textbf{Gaussian Kernels.} We deem the video frames as a series of heatmaps with the queried object as the center of the heat source, and utilize Gaussian kernels to describe the probability distribution of the object's position as shown in Fig. \ref{figspa}. 

Given annotated boxes $\{ b_t\}^{T-1}_{t=0}$, firstly we linearly map it to the feature map scale $L$. For each re-scaled box $b^{\prime}_t$, we find the center coordinates $(x_t,y_t)$. Then, the heatmap $h_t \in [0,1]^{{L} \times L}$ using Gaussian kernel is given by:
\begin{equation}
{h_t(x,y)} =  \exp\left(-\frac{(x-x_t)^2+(y-y_t)^2}{2\sigma ^2}\right),
\label{gk}
\end{equation}
where $\sigma$ determines the size of kernels.

\noindent\textbf{Point Localization.} In this step, our aim is to learn the predicted heatmaps $\{\hat{h}_t\}^{T-1}_{t=0}$ supervised by the Gaussian kernel-based one. 
The peak of the Gaussian distribution of key points is regarded as the positive sample while other pixels are regarded as the negative sample. We modify focal loss \cite{lin2017focal} as:
\begin{equation}
{\cal L}_{loc}^{s} = \frac{-1}{M_{foc}}\sum_{txy}
\begin{cases} 
(1-\hat{h}_{txy})^\alpha\log(\hat{h}_{txy}),\ \  \text{if }h_{txy}>\gamma \\
(1-{h}_{txy})^\beta\hat{h}_{txy}^\alpha\log(1-\hat{h}_{txy}),  \text{else}
\end{cases}
\end{equation}
where $M_{foc}$ stands for the number of annotated boxes, $\alpha$ and $\beta$ are hyper-parameters of the focal loss, $\gamma$ is our modified hyper-parameter which determines the number of positive samples.

\noindent\textbf{Size Regression.} Then, a size regression head is employed to define the object size. Each pixel in the annotation box is treated as a regression sample. 
Given the predicted distances $\{\hat{s}_t\}_{t=0}^{T-1} \in \mathbb{R}^{T\times4\times L \times L}$ and ground truth  $\{{s}_t\}$ , we decode the predicted boxes $\{\hat{b}_t\}$ and corresponding annotated boxes $\{{b}_t\}$. Then, GIoU is used as loss for bounding box regression: 
\begin{equation}
{\cal L}_{reg}^s = \frac{1}{M_{iou}}\sum_{(t,x,y)\in a_t}{\rm GIoU}(\hat{b}_{txy},{b}_t),
\label{gio}
\end{equation}
where $M_{iou}$ represents the number of regression samples, i.e., the number of the pixel $(t,x,y)$ in annotation area $a_t$.

\vspace{-5pt}
\subsection{Temporal Location Head}
\begin{figure}[t]
\centering
\includegraphics[width=0.85\columnwidth]{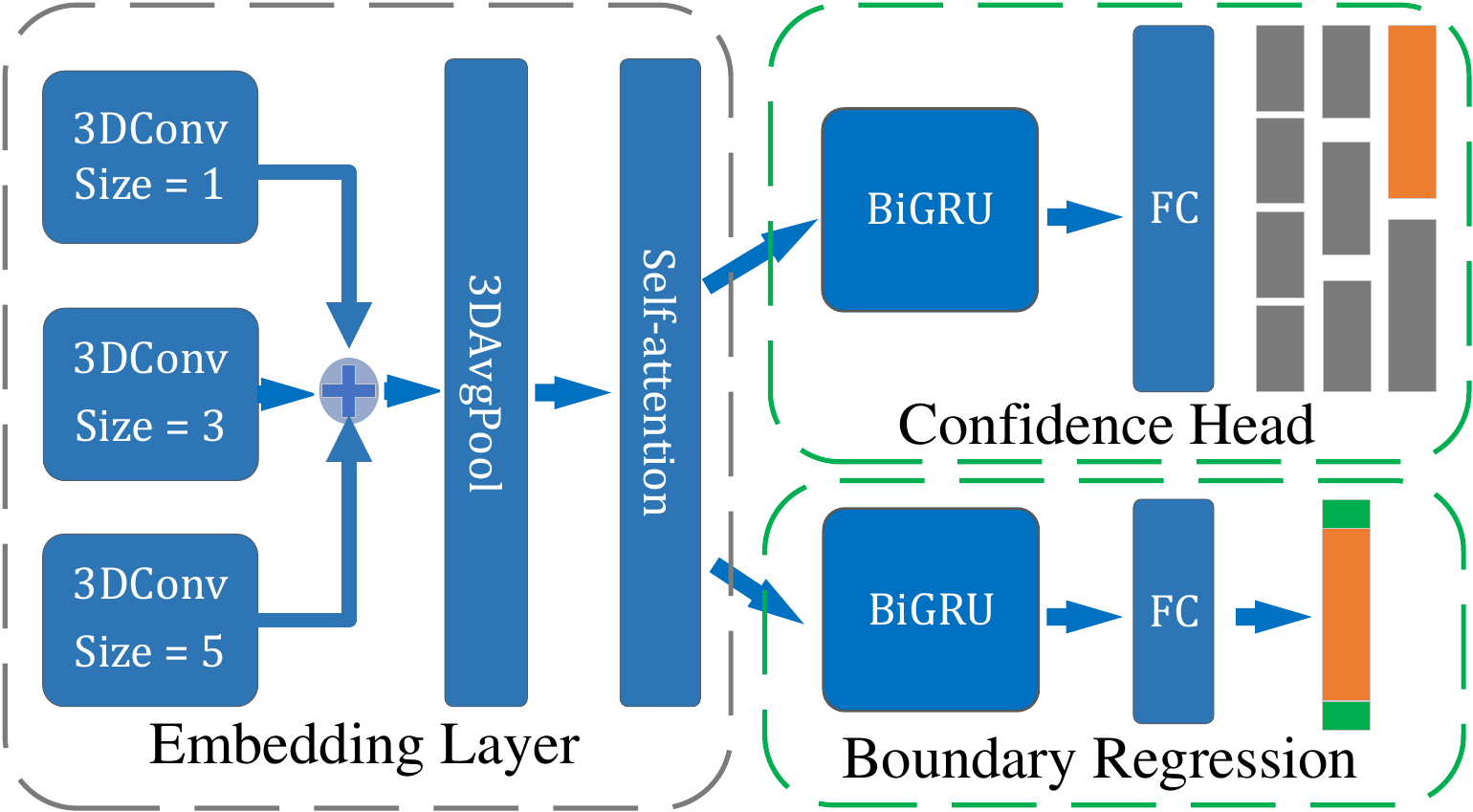}
\vspace{-3pt}
\caption{Illustration of temporal location head.}
\label{figtem}
\vspace{-8pt}
\end{figure}

\vspace{-3pt}
Taking the multi-modal features ${\cal M}_{mix}$ as input, we put it to a temporal location head to attain the temporal boundaries as is shown in Fig. \ref{figtem}.

\noindent\textbf{Embedding Layer.} First, three different kinds of 3D convolution layers are deployed with the kernel size 1, 3 and 5 respectively to learn differentiated time-length features. Then a 3D convolution layer and an average pooling layer are placed after the above layers. Next, we use a self-attention module to enhance the inner relation in terms of time sequence.

\noindent\textbf{Score Confidence Head.} This head is implemented as a
Bi-GRU and a 1D convolution layer. We estimate the IoU $i \in [0,1]$ between the generated temporal tubes and the corresponding ground truth, and confidence scores are the value of the IoUs. Then a threshold $c$ is defined to set the score of the tubes to zero where $i < c$. We utilize a smooth $l_1$ loss for confidence evaluation, given by:

\begin{equation}
{\cal L}_{con}^{t} = \frac{1}{N_c}\sum_{n=1}^{N_c}{\cal L}_1(i_n,\hat{i}_n),
\end{equation}
where ${\cal L}_1$ is the smooth $l_1$ loss, $N_c$ and $\hat{i}_n$ stand for the number of tubes and the $n$-th predicted IoU score.

\noindent\textbf{Boundary Regression Head.} Similarly, we use a
Bi-GRU and a 1D convolution layer to implement the boundary regression head. Every tube that has the potential to be selected own an offset $\delta_k = (\delta_s, \delta_e)$. With the ground truth $(t_s,t_e)$ and the predicted tube $(\hat{t}_s,\hat{t}_e)$, we know that $\delta_s = t_s - \hat{t}_s$ and $\delta_e = t_e - \hat{t}_e$. Then we compute the smooth $l_1$ distance by:

\begin{equation}
{\cal L}_{reg}^{t} =  \frac{1}{N_c}\sum_{k=1}^{N_c}{\cal L}_1(\delta_k. \hat{\delta}_k).
\end{equation}

Therefore, the total loss is a multiple loss combing the above four loss functions as:

\begin{equation}
{\cal L}=  \alpha_1{\cal L}_{loc}^s+\alpha_2{\cal L}_{reg}^s+\alpha_3{\cal L}_{con}^t+\alpha_4{\cal L}_{reg}^t,
\end{equation}
where $\alpha_1,\alpha_2,\alpha_3,\alpha_4$ are balanced parameters for loss.
\vspace{-3pt}
\section{Experiments}
\vspace{-3pt}

\begin{table}[t]

\centering

\begin{tabular}{ccccc}

\hline
Anchor-based & m\_t                  & m\_v & v@0.3 & v@0.5 \\
\hline\hline
G \cite{rohrbach2016grounding} + T \cite{gao2017tall} & \multirow{3}{*}{34.63 } & 9.78   & 11.04   & 4.09    \\
S \cite{yamaguchi2017spatio} + T \cite{gao2017tall}      &                          & 10.40  & 12.38   & 4.27    \\
W \cite{chen2019weakly} + T \cite{gao2017tall}      &                          & 11.36  & 14.63   & 5.91    \\
\cline{1-5}
G \cite{rohrbach2016grounding} + L \cite{chen2019localizing} & \multirow{3}{*}{40.86 } & 11.89  & 15.32   & 5.45    \\
S \cite{yamaguchi2017spatio} + L \cite{chen2019localizing}       &                          & 12.93  & 16.27   & 5.68    \\
W \cite{chen2019weakly} + L \cite{chen2019localizing}     &                          & 14.45  & 18.00   & 7.89    \\
\cline{1-5}
STGRN \cite{zhang2020does}            & 48.47                   & \textbf{19.75} & \textbf{25.77}  & \textbf{14.60}  \\
\hline
\hline
Anchor-free   & & & & \\
\hline
\hline
BM  & 32.62  &   7.21  & 9.03   & 3.87   \\
GKCMN       &     \textbf{55.01}             & \textbf{19.26}     & \textbf{23.53}    & \textbf{15.41}  \\
\hline
\end{tabular}
\vspace{-3pt}
\caption{Performance compared with previous methods on the VidSTG dataset. The letter G, S, W, T, L represents the method of GroundR, STPR, WSSTG, TALL, L-Net respectively}
\vspace{-10pt}
\label{table1}
\end{table}
\noindent\textbf{Dataset.} We evaluate our method on a large-scale spatio-temporal video grounding dataset VidSTG \cite{zhang2020does}, which contains 5,563, 618 and 743 untrimmed videos in the training, validation and testing sets respectively.

\noindent\textbf{Evaluation Metric.} Following previous work \cite{zhang2020does}, we adopt m\_tIoU (m\_t) to evaluate the temporal grounding performance and apply m\_vIoU (m\_v) and vIoU@R (v@R) as the evaluation criteria of spatio-temporal accuracy. 

\noindent\textbf{Implementation Details.}
For video preprocessing, we first resize the input video to $224 \times 224$ pixels for each frame, and put every 8 frames to a pre-trained I3D model with stride 4, obtaining the feature of the last convolutional layer with size $7 \times 7$. The feature dimension $d$ is 32. We set the length of video feature sequences to 200. For language encoding, we set the maximum length of words to 20, and apply Glove embedding to embed each word to a 300 dimension feature matrix. As for the model setting, we set the number of attention layers to 2, confidence threshold $c$ is set to 0.3, $\gamma$ is set to 0.8, $\delta$ is set to 0.9, and we set feature maps scale $L$ to 16. The balanced parameters $\alpha_1,\alpha_2,\alpha_3,\alpha_4$ are $1.0,2.0,0.2,0.1$, respectively. We use an Adam optimizer with a learning rate of 0.003. 

\noindent\textbf{Experiments Results.} 
We remove our mixed convolutional network and location heads to train the model as our anchor-free baseline method (BM). Besides, we compare our GKCMN with the state-of-the-art anchor-based method STGRN \cite{zhang2020does}. Other anchor-based methods like GroundeR (G) \cite{rohrbach2016grounding}, STPR (S) \cite{yamaguchi2017spatio}, WSSTG (W) \cite{chen2019weakly} need a temporal localizer, such as TALL (T) \cite{gao2017tall} and L-Net (L) \cite{chen2019localizing}, to ground spatio-temporal tube in untrimmed videos. Thus, we combined them into six additional methods for comparison.

Table \ref{table1} shows the overall results of experiments and we find: for temporal grounding, our GKCMN outperforms the anchor-based methods STGRN, TALL and L-Net on metric m\_tIoU, which demonstrates the mixed spaio-temporal modeling is of vital importance to capture the temporal characteristics. For spatio-temporal grounding, our model has shown comparable performance with the state-of-the-art anchor-based method, even surpassing STGRN with respect to v@0.5. In the aspect of anchor-free methods, our GKCMN shows a significant improvement over the anchor-free BM.

\begin{table}[t]
\centering
\begin{tabular}{cccc}
\hline
Method            & m\_v & v@0.3 & v@0.5 \\ \hline\hline
GroundeR \cite{rohrbach2016grounding}            & 28.80  & 43.20   & 22.74   \\
STPR \cite{yamaguchi2017spatio}                 & 29.72  & 44.78   & 23.83   \\
WSSTG \cite{chen2019weakly}               & 33.32  & 50.01   & 29.98   \\ 
STGRN \cite{zhang2020does}               &  \textbf {38.04}& \textbf {54.47}  & \textbf{34.80}  \\ \hline\hline
BM                   & 22.37  & 34.92   & 16.33   \\ 
GKCMN      & \textbf {37.25} &\textbf {52.67} &\textbf{34.51} \\ \hline

\end{tabular}
\vspace{-3pt}
\caption{Evaluation results with the temporal ground truth.}
\label{table2}

\end{table}

In order to eliminate the influence of temporal grounding on the overall spatio-temporal grounding, we conducted a separate experiment by giving the temporal ground truth as shows in Table \ref{table2}. Here, we can clearly observe that the proposed GKCMN outperforms the other anchor-based methods and achieves a comparable result with STGRN. It is worth noting that our Gaussian kernel design has greatly improved the accuracy of spatial localization compared with BM.

\begin{table}[t]
\centering
{ 
\begin{tabular}{ccccc}
\hline 
Module  & m\_t             & m\_v & v@0.3&v@0.5\\ \hline\hline
w/o SC  &    54.16            & 18.76  & 22.35   & 13.86   \\
w/o PC  &    54.22            & 18.62  & 22.17   & 13.28   \\
w/o MN  &    53.27            & 17.28  & 21.59   & 12.86   \\
w/o TA  &    53.13            & 18.88  & 22.81   & 14.73   \\ 
w/o SK  &    54.95           & 18.29  & 22.68   & 14.27     \\ \hline
full & \textbf {55.01} & \textbf {19.26} &\textbf {23.53} &\textbf  {15.41} \\ \hline

\end{tabular}
}
\vspace{-3pt}
\caption{Ablation study on the VidSTG dataset.
}
\vspace{-10pt}
\label{table3}
\end{table}

\noindent \textbf{Ablation Study.} For ablation study, we verify the contribution of each part of our proposed GKCMN with the center-based scheme. More specifically, we modify our complete model to five settings: w/o SC, w/o PC, w/o MN, w/o TA, w/o SK, which represents the removal of the serial connection, the parallel connection, the complete mixed convolutional network, the self-attention module, and the replacement Gaussian kernel with single point localization, respectively.


The ablation results are shown in Table \ref{table3}. We can find that every ablation model has precision reduction compared with the full model, which manifests each above component provides a positive contribution.


\vspace{-3pt}
\section{Conclusion}
\label{sec:majhead}
\vspace{-3pt}
In this paper, we propose a novel anchor-free cross-modal network GKCMN for STVG task. The main contributions of our work are: 1) we propose a Gaussian kernel-based anchor-free architecture for STVG task, 2) we develop a mixed convolutional network to capture cross-modal features in both temporal and spatial aspects, 3) experimental results on VidSTG dataset show the superiority of our method.

\noindent \textbf{Acknowledgments.} This work was supported by National Natural Science Foundation of China (NSFC) under No. 61972448.

\bibliographystyle{IEEEbib}

\bibliography{strings,refs}

\begin{thebibliography}{10}

\bibitem{hu2016natural}
Ronghang Hu, Huazhe Xu, Marcus Rohrbach, Jiashi Feng, Kate Saenko, and Trevor
  Darrell,
\newblock ``Natural language object retrieval,''
\newblock in {\em Proceedings of the IEEE Conference on Computer Vision and
  Pattern Recognition}, 2016, pp. 4555--4564.

\bibitem{yu2018mattnet}
Licheng Yu, Zhe Lin, Xiaohui Shen, Jimei Yang, Xin Lu, Mohit Bansal, and
  Tamara~L Berg,
\newblock ``Mattnet: Modular attention network for referring expression
  comprehension,''
\newblock in {\em Proceedings of the IEEE Conference on Computer Vision and
  Pattern Recognition}, 2018, pp. 1307--1315.

\bibitem{yang2019cross}
Sibei Yang, Guanbin Li, and Yizhou Yu,
\newblock ``Cross-modal relationship inference for grounding referring
  expressions,''
\newblock in {\em Proceedings of the IEEE/CVF Conference on Computer Vision and
  Pattern Recognition}, 2019, pp. 4145--4154.

\bibitem{yang2019dynamic}
Sibei Yang, Guanbin Li, and Yizhou Yu,
\newblock ``Dynamic graph attention for referring expression comprehension,''
\newblock in {\em Proceedings of the IEEE/CVF International Conference on
  Computer Vision}, 2019, pp. 4644--4653.

\bibitem{gao2017tall}
Jiyang Gao, Chen Sun, Zhenheng Yang, and Ram Nevatia,
\newblock ``Tall: Temporal activity localization via language query,''
\newblock in {\em Proceedings of the IEEE international conference on computer
  vision}, 2017, pp. 5267--5275.

\bibitem{anne2017localizing}
Lisa Anne~Hendricks, Oliver Wang, Eli Shechtman, Josef Sivic, Trevor Darrell,
  and Bryan Russell,
\newblock ``Localizing moments in video with natural language,''
\newblock in {\em Proceedings of the IEEE international conference on computer
  vision}, 2017, pp. 5803--5812.

\bibitem{liu2020jointly}
Daizong Liu, Xiaoye Qu, Xiao-Yang Liu, Jianfeng Dong, Pan Zhou, and Zichuan Xu,
\newblock ``Jointly cross-and self-modal graph attention network for
  query-based moment localization,''
\newblock in {\em Proceedings of the 28th ACM International Conference on
  Multimedia}, 2020, pp. 4070--4078.

\bibitem{liu2021context}
Daizong Liu, Xiaoye Qu, Jianfeng Dong, Pan Zhou, Yu~Cheng, Wei Wei, Zichuan Xu,
  and Yulai Xie,
\newblock ``Context-aware biaffine localizing network for temporal sentence
  grounding,''
\newblock in {\em Proceedings of the IEEE/CVF Conference on Computer Vision and
  Pattern Recognition}, 2021.

\bibitem{zhang2020does}
Zhu Zhang, Zhou Zhao, Yang Zhao, Qi~Wang, Huasheng Liu, and Lianli Gao,
\newblock ``Where does it exist: Spatio-temporal video grounding for multi-form
  sentences,''
\newblock in {\em Proceedings of the IEEE/CVF Conference on Computer Vision and
  Pattern Recognition}, 2020, pp. 10668--10677.

\bibitem{zhou2018weakly}
Luowei Zhou, Nathan Louis, and Jason~J Corso,
\newblock ``Weakly-supervised video object grounding from text by loss
  weighting and object interaction,''
\newblock {\em arXiv preprint arXiv:1805.02834}, 2018.

\bibitem{vasudevan2018object}
Arun~Balajee Vasudevan, Dengxin Dai, and Luc Van~Gool,
\newblock ``Object referring in videos with language and human gaze,''
\newblock in {\em Proceedings of the IEEE Conference on Computer Vision and
  Pattern Recognition}, 2018, pp. 4129--4138.

\bibitem{yamaguchi2017spatio}
Masataka Yamaguchi, Kuniaki Saito, Yoshitaka Ushiku, and Tatsuya Harada,
\newblock ``Spatio-temporal person retrieval via natural language queries,''
\newblock in {\em Proceedings of the IEEE International Conference on Computer
  Vision}, 2017, pp. 1453--1462.

\bibitem{carreira2017quo}
Joao Carreira and Andrew Zisserman,
\newblock ``Quo vadis, action recognition? a new model and the kinetics
  dataset,''
\newblock in {\em proceedings of the IEEE Conference on Computer Vision and
  Pattern Recognition}, 2017, pp. 6299--6308.

\bibitem{pennington2014glove}
Jeffrey Pennington, Richard Socher, and Christopher~D Manning,
\newblock ``Glove: Global vectors for word representation,''
\newblock in {\em Proceedings of the 2014 conference on empirical methods in
  natural language processing (EMNLP)}, 2014, pp. 1532--1543.

\bibitem{vaswani2017attention}
Ashish Vaswani, Noam Shazeer, Niki Parmar, Jakob Uszkoreit, Llion Jones,
  Aidan~N Gomez, {\L}ukasz Kaiser, and Illia Polosukhin,
\newblock ``Attention is all you need,''
\newblock in {\em Advances in neural information processing systems}, 2017, pp.
  5998--6008.

\bibitem{chung2014empirical}
Junyoung Chung, Caglar Gulcehre, KyungHyun Cho, and Yoshua Bengio,
\newblock ``Empirical evaluation of gated recurrent neural networks on sequence
  modeling,''
\newblock {\em arXiv preprint arXiv:1412.3555}, 2014.

\bibitem{lin2017focal}
Tsung-Yi Lin, Priya Goyal, Ross Girshick, Kaiming He, and Piotr Doll{\'a}r,
\newblock ``Focal loss for dense object detection,''
\newblock in {\em Proceedings of the IEEE international conference on computer
  vision}, 2017, pp. 2980--2988.

\bibitem{rohrbach2016grounding}
Anna Rohrbach, Marcus Rohrbach, Ronghang Hu, Trevor Darrell, and Bernt Schiele,
\newblock ``Grounding of textual phrases in images by reconstruction,''
\newblock in {\em European Conference on Computer Vision}. Springer, 2016, pp.
  817--834.

\bibitem{chen2019weakly}
Zhenfang Chen, Lin Ma, Wenhan Luo, and Kwan-Yee~K Wong,
\newblock ``Weakly-supervised spatio-temporally grounding natural sentence in
  video,''
\newblock {\em arXiv preprint arXiv:1906.02549}, 2019.

\bibitem{chen2019localizing}
Jingyuan Chen, Lin Ma, Xinpeng Chen, Zequn Jie, and Jiebo Luo,
\newblock ``Localizing natural language in videos,''
\newblock in {\em Proceedings of the AAAI Conference on Artificial
  Intelligence}, 2019, vol.~33, pp. 8175--8182.

\end{thebibliography}

\end{document}